\pdfoutput=1 
\documentclass[preprint,3p,12pt,authoryear]{elsarticle} 

\usepackage[utf8]{inputenc} 
\usepackage[T1]{fontenc}    
\usepackage{hyperref}       
\usepackage{url}            
\usepackage{booktabs}       
\usepackage{amsfonts}       
\usepackage{nicefrac}       
\usepackage{microtype}      
\usepackage{lipsum}
\usepackage{graphicx}
\usepackage{physics}
\usepackage{amsmath}
\usepackage{wrapfig}
\usepackage{algorithm}
\usepackage{algorithmic}

\journal{arXiv}

\begin{document}

\begin{frontmatter}

\title{Fading memory as inductive bias in residual recurrent networks}

\author[inst1,inst2]{Igor Dubinin}
\ead{igor.dubinin@esi-frankfurt.de}

\author[inst1]{Felix Effenberger}
\ead{felix.effenberger@esi-frankfurt.de}


\affiliation[inst1]{organization={Ernst Strüngmann Institute},
            addressline={Deutschordenstraße 46}, 
            city={Frankfurt am Main},
            postcode={60528}, 
            country={Germany}}

\affiliation[inst2]{organization={Frankfurt Institute for Advanced Studies},
            addressline={Ruth-Moufang-Straße 1}, 
            city={Frankfurt am Main},
            postcode={60438}, 
            country={Germany}}

\begin{abstract}
Residual connections have been proposed as an architecture-based inductive bias to mitigate the problem of exploding and vanishing gradients and increased task performance in both feed-forward and recurrent networks (RNNs) when trained with the backpropagation algorithm.
Yet, little is known about how residual connections in RNNs influence their dynamics and fading memory properties.
Here, we introduce weakly coupled residual recurrent networks (WCRNNs) in which residual connections result in well-defined Lyapunov exponents and allow for studying properties of fading memory.
We investigate how the residual connections of WCRNNs influence their performance, network dynamics, and memory properties on a set of benchmark tasks.
We show that several distinct forms of residual connections yield effective inductive biases that result in increased network expressivity.
In particular, those are residual connections that (i) result in network dynamics at the proximity of the edge of chaos, (ii) allow networks to capitalize on characteristic spectral properties of the data, and (iii) result in heterogeneous memory properties.
In addition, we demonstrate how our results can be extended to non-linear residuals and introduce a weakly coupled residual initialization scheme that can be used for Elman RNNs.

\end{abstract}


\end{frontmatter}

\section{Introduction}
\noindent
The power of artificial neural networks in solving tasks lies in their universal approximation abilities, which is commonly referred to as \emph{theoretical expressivity}~\citep{hornik1989multilayer,cybenko1989approximation,funahashi1989approximate,barron1994approximation}. 
However, the practical solutions to which networks can converge in a reasonable number of training iterations of a typically gradient-based learning scheme such as backpropagation, the \emph{practical expressivity} of a network, have been shown to lag behind their theoretical expressivity~\citep{hanin2019complexity}.
Practical expressivity is determined by a set of inductive biases that take the form of (i) network architecture, (ii) weight initialization methods, (iii) convergence properties and other specifics of the training procedure, and (iv) more generally, comprise anything that influences the space of mappings learnable by a given network in practice~\citep{battaglia2018relational,goyal2022inductive}.
As the bias-variance trade-off suggests, the right choice of inductive biases plays a crucial role for model performance because properly informed biases can improve the efficiency of learning, a serious constraint on network expressivity in practice~\citep{kearns1994introduction}.

A celebrated example of feed-forward networks with an effective inductive bias in the form of a constrained network architecture are convolutional neural networks~\citep{lecun1995convolutional}.
Another popular form of an architectural inductive bias are so-called \emph{residual connections} (or skip connections) of deep feed-forward architectures.
These have been shown to strongly increase performance for many architectures and are used, for example, in the U-Net~\citep{ronneberger2015u}, ResNet~\citep{he2016deep} or Transformer~\citep{vaswani2017attention} architectures.
Such residual connections have been shown to prevent gradients from vanishing in deep feed-forward networks, thereby mitigating one aspect of the well-studied exploding and vanishing gradients problem (EVGP) that appears in practice when training deep networks with the backpropagation algorithm~\citep{glorot2010understanding}.

When considering inductive biases in recurrent neural networks (RNNs), the crucial question is how these biases influence network dynamics and the resulting memory properties of the networks.
Training an RNN with the backpropagation through time (BPTT) algorithm involves unrolling the RNN into a deep feed-forward network, so that networks trained on longer time series also face the EVGP~\citep{pascanu2013difficulty}.
Historically, RNNs have been studied from a dynamical systems perspective~\citep{bengio1994learning,hochreiter1997long} and many ideas have been proposed to address the EVGP~\citep{schoenholz2016deep, chang2019antisymmetricRNN, miller2018stable, erichson2020lipschitz}.
Importantly, the dynamical systems approach showed that RNN dynamics which are close to the point of a transition between stability and instability (the edge of chaos) are characterized by long-term fading memory and therefore efficient gradient propagation~\citep{vogt2020lyapunov,engelken2020lyapunov}. 

Although the influence of residual connections on RNN performance has been studied previously~\citep{wang2016recurrent, yue2018residual}, dynamics and memory properties of RNNs with residual connections have not been studied in detail.
Here, we fill this gap and explore how residual connections in RNNs can result in inductive biases that influence the networks' dynamics and properties of their fading memory.
Note that the present study was not primarily motivated by the goal of developing a model that achieves a new state of the art (SOTA) score in a number of benchmark tasks, but to study RNN dynamics and fading memory properties by means of Lyapunov exponents.
The main contributions of this work are as follows.

\begin{itemize} 

\item We extend on the connection between network dynamics, fading memory, and learning dynamics in RNNs discussed in \cite{pascanu2013difficulty}, showing that the fading memory properties of RNN dynamics result in temporally modulated learning rates.

\item
A new RNN architecture, the \emph{weakly coupled residual recurrent network} (WCRNN), is introduced and proven to have stable and easily controllable memory properties by showing the existence of Lyapunov exponents of network dynamics. In particular, we demonstrate how the eigenvalues of the residual matrix control fading memory in WCRNNs. Additionally, we study WCRNNs with dynamics close to the edge of chaos and confirm the theoretically predicted trade-off between the efficiency (i.e. in how many training steps the network converges to a high-performing configuration) and the stability of learning.

\item We show that informed residual connections and corresponding inductive biases result in higher practical expressivity of WCRNNs on a set of benchmark problems, assessed by learning efficiency and the best test accuracy achieved. In particular, we show how residuals resulting in weakly subcritical network dynamics allow the networks to benefit from long memory timescales, how residuals with rotational dynamics allow the networks to utilize spectral properties of the data samples, and how heterogeneous residuals allow the networks to capitalize on the resulting diversity of informed memory timescales.

\item We show how results from WCRNNs with linear residuals can be generalized to the case of non-linear residuals and to the general case of a standard Elman RNN in the form of a weakly coupled residual initialization scheme.

\end{itemize}

\noindent
In summary, our work demonstrates how Lyapunov exponents can be used to characterize fading memory resulting from residual connections and shows how informed residual connections can be used to achieve superior practical expressivity in RNNs.

\section{Background and motivation}

\noindent
In this section, we discuss the connection between network dynamics, memory, and learning dynamics resulting from the training by backpropagation through time (BPTT). In Section \ref{ss:analysis}, we introduce the concept of Lyapunov exponents and show how they determine memory timescales. In Section \ref{ss:learning}, we show how learning dynamics, mediated by BPTT, is influenced by the properties of fading memory.

\subsection{Dynamical systems analysis}
\label{ss:analysis}
The memory of a system, commonly called fading memory in the context of recurrent neural networks, is a well-defined and thoroughly studied concept in the field of dynamical systems. 
From a dynamical system's perspective, a recurrent network is a non-autonomous, non-linear recurrent discrete map, the dynamics of which are given by
\begin{equation}
\mathbf{x}_{t+1}=\mathbf{f}(\mathbf{x}_{t}, \mathbf{S}_{t}),
\label{eq:disc_map}
\end{equation}
where $\mathbf{x}_{t} \in \mathbb{R}^N$ is the network state at time $t$, $\mathbf{f}: \mathbb{R}^N\rightarrow \mathbb{R}^N$  is non-linear function, $\mathbf{S}_{t} \in \mathbb{R}^N$ is the input to the network at time $t$ and $N$ is the dimensionality of the network state. 

First, we consider the autonomous case, where the memory properties of network dynamics can be studied by means of perturbation theory and Lyapunov exponents. 
If we evolve an infinitesimal perturbation of the $P$-dimensional volume of the tangent space $\delta \mathbf{P}$, linearize it along this perturbation and apply the chain rule $t$ times, we obtain
\begin{equation}
\delta \mathbf{P}(t+1) = \mathbf{V}_{\mathbf{x}}(f^t(\mathbf{x}_1)) \delta \mathbf{P}(1),
\label{eq:var_general_vol}
\end{equation}
where $\mathbf{x}_1$ and $\delta \mathbf{P}(1)$ are the initial state and initial volume of the tangent space, $\mathbf{f}^t=\mathbf{f} \circ \dots \circ \mathbf{f}$ denotes the $t$-fold iteration of the map $\mathbf{f}$, and $\mathbf{V}_{x}(f^t(\mathbf{x}_1))$ denotes a variational term. 
In the case of a discrete map $\mathbf{f}$, the variational term $\mathbf{V}_{x}(f^t(\mathbf{x}_1))$ takes the form of the product of the instantaneous Jacobians $\mathbf{J}_x$ of $\mathbf{f}$ over the course of the system trajectory~\citep{eckmann1985ergodic,sandri1996numerical} and can be written as 
\begin{equation}
\mathbf{V}_{\mathbf{x}}(f^t(\mathbf{x}_1)) = \mathbf{J}_{\mathbf{x}}(f^{t}(\mathbf{x}_t))\mathbf{J}_{\mathbf{x}}(f^{t-1}(\mathbf{x}_{t-1}))...\mathbf{J}_{\mathbf{x}}(f(\mathbf{x}_1)).
\label{eq:var_disc_lin_eq}
\end{equation}
Finally, according to Oseledets' theorem~\citep{oseledets1968multiplicative}, the Lyapunov exponents for the autonomous system (\ref{eq:disc_map}) are given by the eigenvalues of the matrix
\begin{equation}
\mathbf{M}=\lim_{t \to \infty} \frac{1}{2t}\log(\mathbf{V}_x(f^t(\mathbf{x}_1)) \mathbf{V}^T_x(f^t(\mathbf{x}_1))),
\label{eq:ose}
\end{equation}
where $\cdot^T$ indicates the matrix transpose. 

Essentially, the Jacobians $\mathbf{J}_{x}(f^{k})$ define the space of instantaneous local volume transformations that are accessible to the system. 
The variational term $\mathbf{V}_{x}(f^t)$ determines the memory timescales of network dynamics, and every Lyapunov exponent defines the direction with an asymptotically stable rate of memory change. 
The number of Lyapunov exponents is equal to the dimensionality of the system $N$ and
the largest Lyapunov exponent determines the stability of network dynamics, with a negative (positive) Lyapunov exponent indicating its stability (instability). 
In particular, the largest Lyapunov exponent changes its sign when the system undergoes a transition between chaos and order. 
 
In the general case given in (\ref{eq:disc_map}), the external input makes the system non-autonomous and the existence of the limit (\ref{eq:ose}) is not guaranteed. 
Thus, the described analysis cannot be easily performed for most input-driven recurrent networks.
In this study, we show how the introduction of weak coupling can mitigate this issue; see Section \ref{s:resid}.  
 
\subsection{Learning dynamics}
\label{ss:learning}

Through the lens of the theory of dynamical systems, training an RNN of the form (\ref{eq:disc_map}) with backpropagation through time creates the following learning dynamics $\mathbf{g}$ on the recurrent weights $\mathbf{w}$, 
\begin{equation}
\mathbf{w}_{\tau+1} = \mathbf{g}(\mathbf{w}_{\tau}) = \mathbf{w}_{\tau} - \eta \frac{1}{M} \sum_{M}\nabla_{w}L,
\label{eq:learn_dyn}
\end{equation}
where $\tau\in \mathbb{N}$ denotes the training iteration, $\eta\in\mathbb{R}^{+}$ is the learning rate hyperparameter, $\nabla_{w}L$ is the gradient of the loss function $L$  with respect to the weight $w$, and $M\in\mathbb{N}$ is the batch size. 
Here, we assume a deterministic version of gradient descent without loss of generality. 
In practice, training an RNN with BPTT involves unrolling the network over time so that the RNN is transformed into an equivalent feed forward network consisting of $D$ \emph{unrolled recurrent layers} (one for each time point $1\leq t\leq D$), and also defines a temporal distance between any two unrolled recurrent layers.
To obtain network predictions, all networks are equipped with an affine readout layer that transforms the state vector of the network at the last time point $t=D$ into activations of a set of output units on which the loss function $L$ is computed (see Fig.~\ref{fig:scheme}).

The theoretical considerations derived in this section can be applied to all typical loss functions used for classification and regression problems, as long as the Hessian of the loss function with respect to network predictions is positive semidefinite and its derivative at the optimum has a vanishing mean~\cite{schraudolph2002fast}.
In our experiments, we used loss functions for which these conditions are met, namely a cross-entropy loss with softmax for classification problems and a root-mean-square (RMS) loss for regression tasks. 
For a detailed description of the experiments performed, see Section~\ref{s:exps}.

The instantaneous Jacobian $\mathbf{J}_{w}$ of the learning dynamics (\ref{eq:learn_dyn}) is then given by
\begin{equation}
\mathbf{J}_{w}(\mathbf{g})=\mathbf{I}-\eta\frac{1}{M}\sum_{M}\mathbf{H}_{w}(L), 
\label{eq:jac_learn}
\end{equation}
where $\mathbf{H}_{w}(L)$ denotes the Hessian of the loss function $L$ and $\mathbf{I}$ is the identity matrix.
Note that the asymptotic behavior of the product of instantaneous Jacobians given in (\ref{eq:jac_learn}) determines the convergence or divergence of the learning dynamics, in the same way as the variational term $\mathbf{V}_{x}$ determines the convergence of divergence of the recurrent network dynamics in (\ref{eq:disc_map}).
This agrees with previous studies that have shown that the curvature of the loss landscape defined by the Hessian plays a crucial role in gradient-based learning~\citep{dauphin2014identifying}.

From (\ref{eq:jac_learn}), it also follows that every eigenvalue of the Hessian $\mathbf{H}_{w}(L)$ defines an \emph{effective learning rate} in the direction of the associated eigenvector in the weight space, modulating the base learning rate $\eta$ in the direction of this eigenvector.
This effective learning rate is equal to $(1-\eta \lambda_{\mathbf{H}}^{M})^{-1}$, where $\lambda_{\mathbf{H}}^{M}$ is the corresponding eigenvalue of the Hessian averaged over a given batch.
In order to further investigate these effective learning rates, we can rewrite the Hessian as
\begin{equation}
\mathbf{H}_{w}(L) = \mathbf{J}_{w}^{T}(f^{D}(\mathbf{x}_1)) \mathbf{H}_{f^{D}}(L) \mathbf{J}_{w}(f^{D}(\mathbf{x}_1)) + \sum_{n=1}^{N}\nabla_{f^{D}_{n}}(L)\mathbf{H}_{w}(f^{D}_{n}(\mathbf{x}_1)),
\label{eq:hessian_gen}
\end{equation}
where $D$ denotes the input length, $\mathbf{J}_{w}(f^{D})$ is the Jacobian of $f^{D}$ with respect to $w$, $\nabla_{f^{D}_{n}}(L)$ is the gradient of the loss function $L$ with respect to the $n$-th coordinate of $f^{D}$, and $\mathbf{H}_{w}({f^{D}})_{n}$ is the Hessian of the $n$-th coordinate of $f^{D}$ with respect to $w$, see~\citep{schraudolph2002fast}.
Here, we consider the loss to be a function of the network state at the last time point $L(f^{D})$. 
As therefore the decoding layer is included in $\mathbf{H}_{f^{D}}(L)$, this allows us to directly show the dependence of $\mathbf{H}_{w}(L)$ on $\mathbf{J}_{w}(f^{D})$.

The first term in (\ref{eq:hessian_gen}) is known as the Generalized Gauss-Newton (GGN) matrix, a popular approximation for the curvature matrix in second-order optimization methods~\citep{thomas2020interplay}.
For the loss functions considered here, the Hessian converges to the GGN matrix when training by BPTT because the second term in (\ref{eq:hessian_gen}) is proportional to the loss and vanishes as the learning approaches a local minimum of the loss landscape.
Therefore, the asymptotic behavior of learning dynamics given in (\ref{eq:learn_dyn}) is predominately influenced by the properties of the GGN matrix.

For a recurrent network defined by (\ref{eq:disc_map}), the GGN matrix depends on the Jacobians $\mathbf{J}_{w}(f^{t}(\mathbf{x}_t)))$ and can be computed by applying the chain rule as 
\begin{equation}
\mathbf{J}_{w}(f^{D}(\mathbf{x}_1)) = \sum_{1\leq t \leq D}
\mathbf{V}_{x}^{D-t}(f^{D}(\mathbf{x}_{t+1}))
\mathbf{J}_{w}(f^{t} (\mathbf{x}_t)),
\label{eq:grad_prop}
\end{equation}
where $\mathbf{V}_{x}^{D-t}(f^D(x_{t+1})) = \mathbf{J}_{x}(f^{D}(\mathbf{x}_{D}))\mathbf{J}_{x}(f^{D-1}(\mathbf{x}_{D-1}))\dots \mathbf{J}_{x}(f^{t+1}(\mathbf{x}_{t+1}))$
is a truncated version of the variational term in (\ref{eq:var_disc_lin_eq}), and $\mathbf{V}_{x}^{0} = I$. 
This shows that the memory properties of network dynamics influence the GGN matrix and thereby the final configuration to which the network converges. 
Taken together, RNN dynamics thus plays the role of an inductive bias.

Importantly, the second term in (\ref{eq:hessian_gen}) can be analyzed further if we apply chain rule to the Hessian $t$ times, and we obtain
\begin{equation}
\begin{aligned}
\sum_{n=1}^{N}\nabla_{f^{D}_{n}}(L)\mathbf{H}_{w}({f^{D}_{n}}(\mathbf{x}_1)) =  \sum_{n=1}^{N} \sum_{k=2}^{D} \nabla_{f^{t}_{n}}(L) \mathbf{C}^{f^{t}_{n}} +\sum_{n=1}^{N}\nabla_{f_{n}}(L)\mathbf{H}_{w}({f_{n}}(\mathbf{x}_1)), 
\label{eq:hessian_2term}
\end{aligned}
\end{equation}
where we denote the curvature matrix at time $t$ as $\mathbf{C}^{f^{t}_{n}} = (\mathbf{J}_{w}(f^{t-1}))^{T} \mathbf{H}_{f^{t-1}}(f^{t}_{n}) (\mathbf{J}_{w}(f^{t-1}))$ with total Jacobians evaluated at $\mathbf{x}_1$, and by $\mathbf{H}_{f^{t-1}}({f^{t}_{n}})$ the Hessian of $n$-th coordinate of $f^{t}$ with respect to the previous state $f^{t-1}$, evaluated as instantaneous partial derivatives. 

Due to the presence of the activation function, the curvature matrix  $\mathbf{C}^{f^{t}_{n}}$ is (in contrast to the GGN matrix) not necessarily positive semi-definite, but we note that this does not affect the following conclusions.
It follows from (\ref{eq:hessian_2term}) that we can represent the eigenvalues of the Hessian of the full network as a sum of the contributions from each unrolled recurrent layer as $\lambda_{\mathbf{H}} = \sum_{1\leq t \leq D} \lambda_{\mathbf{H}}^{f^{t}}$.
Importantly, the magnitudes of these contributions are proportional to the corresponding gradient propagation
\begin{equation}
\lambda_{\mathbf{H}}^{f^{t}} \propto \nabla_{f^{t}}(L) = (\mathbf{V}_{x}^{{D}-t}) ^{T} \nabla_{f^{D}}L,
\label{eq:contrib}
\end{equation}
meaning that the variational term $\mathbf{V}_{x}$ determines the contribution of each unrolled layer to the overall effective learning rates.

Taken together, this shows that the variational term $\mathbf{V}_{x}$ not only defines the memory properties of the network dynamics but also temporally modulates the effective learning rates, establishing a connection between fading memory and learning dynamics.
This connection will be important for the further analysis of weakly coupled residual recurrent networks (WCRNNs) as defined below.

\section{Weakly coupled residual recurrent networks} 
\label{s:resid}

\noindent
Residual connections in deep feed-forward networks have been shown to allow for a better backpropagation of errors and are usually implemented by an identity map between subsequent layers~\citep{he2016deep}. 

\begin{figure*}[ht]
    \centering
    \includegraphics{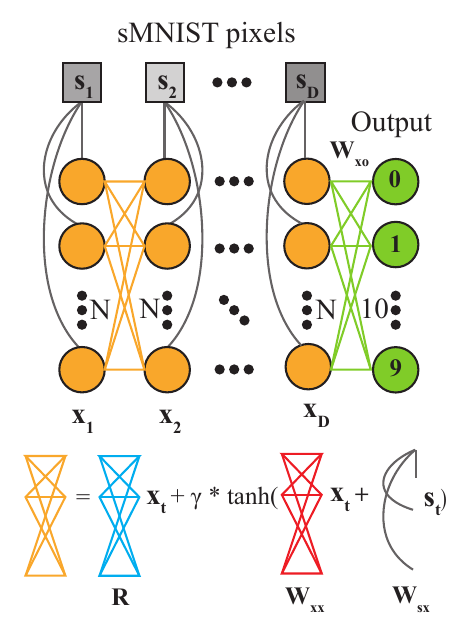}
    \caption{Schematic representation of the WCRNN model for the sMNIST classification task. Each 28x28 pixel MNIST digit is serialized and presented to the network as a time series of length $D=784$. $s_t$ and $x_t$ denote the stimulus and network amplitude configurations at the discrete time step $t$ $(1\leq t \leq 784)$, respectively. Orange circles indicate RNN nodes (unrolled over time) and green circles indicate the output units for the 10 digit classes, respectively. Line colors indicate the input type. The total recurrent input is shown in orange and consists of residual input as mediated by the residual map $R$ shown in blue, the recurrent input as mediated by the recurrent weight matrix $W_{xx}$ shown in red, and the external input mediated by an input projection matrix $W_{sx}$ shown in gray. The readout weights $W_{xo}$ of a linear readout performed at $t=784$ are shown in green. In the case of the ADD datasets, the configuration is analogous, except for adjustments in the input and output layers (2d input, one output unit).
    }
   \label{fig:scheme}
\end{figure*}

Here, we consider the more general case of \emph{weakly coupled residual recurrent neural networks} (WCRNNs) equipped with an arbitrary fixed residual map $\mathbf{R}: \mathbb{R}^N\rightarrow \mathbb{R}^N$. 
The update equation for such networks takes the form
\begin{equation}
\mathbf{x}_{t+1} = \mathbf{R}(\mathbf{x}_t) + \gamma \cdot \sigma(\mathbf{W}_{xx}\mathbf{x}_t + \mathbf{S}_t),
\label{eq:general_residual}
\end{equation}
where  $\gamma\ll1$ denotes a weak coupling constant, $\sigma$ denotes a non-linearity (typically $\tanh$), and $\mathbf{S}_t = \mathbf{W}_{sx} \mathbf{s}_t$ is the input vector that is an affine projection with weights $\mathbf{W}_{sx}$ on some time-varying input data $\mathbf{s}_t$.
We also equip the network with an affine readout layer $\mathbf{W}_{xo}$ and perform the readout on the final network state $\mathbf{x}_{D}$, where $D$ is the length of the input, see Fig.~\ref{fig:scheme}.
The weights $\mathbf{W}_{sx}, \mathbf{W}_{xx}, \mathbf{W}_{xo}$ are subject to backpropagation learning and include trainable bias terms that are omitted in (\ref{eq:general_residual}) for simplicity of notation.

The condition of weak coupling ($\gamma\ll1$) simplifies the analysis of memory properties of such networks because for small values of $\gamma$ the variational term (\ref{eq:var_disc_lin_eq}) can be written as
\begin{equation}
\begin{aligned}
\mathbf{V}_x (f^t(\mathbf{x}_1)) = \mathbf{J}_R(\mathbf{x}_t)\mathbf{J}_R(\mathbf{x}_{t-1})...\mathbf{J}_R(\mathbf{x}_1)+O(\gamma),
\label{eq:var_gen_residual}
\end{aligned}
\end{equation}
where $\mathbf{J}_R(\mathbf{x}_t)$ denotes the instantaneous Jacobian of the residual at time $t$, and $O(\cdot)$ is the Landau O. 

Equation (\ref{eq:var_gen_residual}) shows that the properties of the dynamics of WCRNNs are predominately determined by the instantaneous Jacobians of the residual. 
Thus, the WCRNN architecture allows for a control of the properties of fading memory of the entire network by choosing an appropriate residual.

We note that the weak coupling condition holds as long as $\gamma$ is sufficiently smaller than 1.
Small values of $\gamma$ lead to controllable network dynamics in WCRNNs and are the prerequisite for well-defined Lyapunov exponents, as the analysis below shows.
However, if $\gamma$ becomes negligibly small ($\gamma < 10^{-3}$), the external input will be too small in magnitude to allow the networks to perform well, so there is a trade-off between stabilizing the system dynamics and still allowing for a forcing of system dynamics through an external input.
In practice, we have seen that values of $\gamma\in [0.001, 0.1]$ work well for the datasets tested.

In the general case (\ref{eq:general_residual}), the Lyapunov exponents depend on $x_t$ due to the non-linearity in the residual, which can complicate proving the existence of the limit in (\ref{eq:ose}).
To simplify the analysis, we first consider networks with linear residuals $\mathbf{R}(\mathbf{x}_t) =  \mathbf{R} \mathbf{x}_t$, where $\mathbf{R}$ is a $N \times N$ matrix. 
For the linear case, the instantaneous eigenvalues of the residual are independent of the trajectory of the system, and the variational term simplifies to $\mathbf{V}(f^t(\mathbf{x}_1)) = \mathbf{R}^t+\mathcal{O}(\gamma)$, where $\cdot^{t}$ denotes matrix exponentiation.
Furthermore, we can easily derive the Lyapunov exponents in this case, because the Oseledets equation (\ref{eq:ose}) in the limit of weak coupling yields $\mathbf{M}=\log{\mathbf{A}} + \mathcal{O}(\gamma)$,
where $\mathbf{A}=\operatorname{diag}(\lambda_{1},\lambda_{2},...,\lambda_{N})$ is the diagonal matrix of eigenvalues of $\mathbf{R}$.
This means that for WCRNNs the logarithms of the eigenvalues of the residuals $\lambda_{\text{residual}}$ approximate the Lyapunov exponents $\text{LE}_{\text{net}}$ of the entire network 
\begin{equation}
\text{LE}_{\text{net}} \approx \log{\lambda_{\text{residual}}},
\label{eq:ll_res}
\end{equation}
and that the weak coupling limits the range of their finite-size fluctuations.
Based on their dynamical stability (distance to the edge of chaos, see~\citep{bertschinger2004real}) as determined by the magnitude of the largest eigenvalue $\lambda_{\text{max}}$ of the residual matrix, we can distinguish three classes of WCRNNs: (i) subcritical ($\lambda_{\text{max}}<1$), 
 (ii) critical ($\lambda_{\text{max}} = 1$), and (iii) supercritical ($\lambda_{\text{max}}>1$) networks.

Moreover, it follows from (\ref{eq:res_ggn}) that the Hessian of WCRNN dynamics at the point of convergence takes the form 
\begin{equation}
\mathbf{H}_{w}(L) = \sum_{1\leq k \leq D} \sum_{1\leq m \leq D}
\mathbf{J}_{w}^{T}(f^{k}(\mathbf{x}_k))
(\mathbf{R}^{D-k})^T
\mathbf{H}_{f^{D}}(L) 
\mathbf{R}^{D-m}
\mathbf{J}_{w}(f^{m} (\mathbf{x}_m)) +\mathcal{O}(\gamma^3) .
\label{eq:res_ggn}
\end{equation}
This shows that the contribution of the partial derivatives to the overall curvature at different times is proportional to the corresponding eigenvalues of the residual matrix.
This means that the residual matrices determine the inductive biases in the final weight configuration to which the network converges.
We thus predict that WCRNNs will achieve different levels of performance depending on the residual initialization, and that the residuals resulting in the best performing networks will be dataset-specific.
Moreover, we predict that an optimal residual configuration will depend on the input length.

Similarly to the GGN matrix, the second term of (\ref{eq:grad_prop}) is also affected by the residuals.
It follows from (\ref{eq:contrib}) that
\begin{equation}
\lambda_{\mathbf{H}}^{f^{t}} \propto \nabla_{f^{t}}(L) = (\mathbf{R}^{D-t}) ^T \nabla_{f^{t}}L +\mathcal{O}(\gamma) , 
\label{eq:res_contrib}
\end{equation}
meaning that the magnitudes of eigenvalues of $\mathbf{R}$ determine the contribution of each unrolled recurrent layer to the overall effective learning rates.
Based on the propagation of the gradients as described in (\ref{eq:res_contrib}), we therefore predict temporally modulated effective learning rates, where the eigenvalues of $\mathbf{R}$ define the exponential rate with which this contribution decays or increases with time $t$.
On the one hand, an increase in the magnitudes of the eigenvalues of $\mathbf{R}$ induces an improved learning efficiency for information contained in temporally distant recurrent layers.
However, such an increase can also result in an instability of learning dynamics if the overall effective learning rates reach high magnitudes as a result.
In contrast, a decrease in the magnitudes of the eigenvalues of $\mathbf{R}$ can result in a loss of temporally distant information and reduce the efficiency of learning dynamics.
At the same time, the same mechanism can also provide a better stability of learning dynamics by reducing the overall magnitudes of effective learning rates.

These properties of WCRNNs make us hypothesize that there exists a trade-off between the stability of learning dynamics and \emph{learning efficiency}, the number of iterations required to achieve a certain value of the loss function (with lower numbers of iterations being better).
For more subcritical networks, we expect problems with slow learning and vanishing gradients.
For more supercritical networks, we expect faster convergence, but potentially problems with unstable learning trajectories and exploding gradients at the same time.
For critical WCRNNs, characterized by network dynamics in proximity to the edge of chaos, we anticipate an optimal balance between learning effiency and stability.

In summary, we design weakly coupled residual recurrent networks (WCRNNs), where the introduction of weak coupling allows us to obtain stable memory timescales as defined by Lyapunov exponents.
We show that the properties of fading memory are mainly determined by properties of the residuals and, in the case of linear residuals, by the eigenvalues of the residual matrix.
On the basis of our analyses, we predict that WCRNNs close to criticality show the best trade-off between efficiency and stability of learning, and that optimal residual configurations depend on the input length. 
In conclusion, we have shown here how architecture-based inductive biases in the form of residuals shape the memory properties and learning dynamics of WCRNNs in theory. 
In the next section, we will test our predictions in practice.

\section{Experiments}
\label{s:exps}

\noindent
To empirically validate our theoretical predictions, we performed experiments on several datasets: 

\begin{itemize}
\item
Sequential MNIST (sMNIST), where the 28x28 pixels of MNIST digits~\citep{lecun1998gradient} are presented to the network sequentially in the form of a time series and the task is to solve a digit classification problem. The samples are turned into a time series of length 784 by collecting intensity values in scan-line order from top left to bottom right. We also consider the permuted sequential MNIST (psMNIST) data set, where a random but fixed permutation is applied to the sMNIST samples. The permutation removes the dominant low-frequency components present in the sMNIST samples and increases the difficulty of the classification problem. For these classification tasks, a cross-entropy loss was used. 
\item
The adding problem (ADD) of lengths 100, 200, 400, 800, as a regression problem. Here, every sample is given by a two-dimensional time series of the specified length~\citep{hochreiter1997long}. The first coordinate is given by random numbers drawn from a uniform distribution $[0, 1]$, and the second coordinate constitutes a cue signal taking values $0$ (no cue) and $1$ (cue). The task of the network is to compute the sum of the input values presented in the first coordinate for two cue points randomly placed in the first half and the second half of the signal, respectively. For these tasks, a root mean square (RMS) loss was used. 
\end{itemize}
As the primary goal of this study is not to provide a model outperforming other SOTA architectures, but rather to study network dynamics in WCRNNs and underlying principles of how network dynamics influence learning dynamics and inductive biases, we chose the MNIST dataset and the adding problem for most of our experiments as these are well established classic benchmarks for RNNs. 
These datasets can be made more challenging by introducing permutations for MNIST or longer sample lengths for the adding problem.
To test our models on a more challenging task, we also performed experiments on the gray-scaled sCIFAR10 dataset, which is part of Long Range Arena benchmark designed for long sequence tasks \cite{tay2020long}.
The sCIFAR10 dataset contains 10 different classes of 32x32 pixel images, which are transformed into time series of 1024 gray-scaled pixels, analogously to the sMNIST data set.
As for the sMNIST data set, a cross-entropy loss function was used for the latter.

The experiments were carried out in PyTorch~\citep{paszke2019pytorch} for network sizes of 50, 100, and 200 units.
Training was performed for 200 epochs for the sMNIST and psMNIST datasets and for 150 epochs for the ADD datasets.
Stochastic gradient descent (SGD) with a momentum of $0.9$ was used as an optimizer for BPTT and training iterations were performed according to a minibatch scheme, using batch sizes of 64, 128, and 256 samples.
Qualitatively, results were found to be mostly independent of network and batch size. 
Thus, we present results for networks of 100 units, trained with a batch size of 128 samples in the following.
Results were collected over 5 network instances with random weight initialization, and most plots report mean scores and their standard deviation obtained from these 5 instances.
During initialization, weights and biases were randomly sampled from a Kaiming uniform distribution according to the default implementation of the PyTorch \texttt{torch.\allowbreak nn.\allowbreak Linear} layer ($U(-1/\sqrt{n_{\text{in}}}, 1/\sqrt{n_{\text{in}}})$, where $n_{\text{in}}$ denotes the input dimension of a given layer).  
As the non-linear activation function, we used  $\sigma=\tanh$ in all of our experiments.

In Section \ref{ss:transition} we present simulation results of WCRNNs with dynamics close to the edge of chaos, showing the validity of our theoretical predictions about their performance for all datasets. 
In Sections \ref{ss:rot} and \ref{ss:hete} we show how rotational and heterogeneous residuals can be beneficial to the performance of WCRNNs. 
Lastly, in Section \ref{ss:init} we show that our results can also be generalized to Elman RNNs by an initialization scheme.
The results presented were found to be consistent across all datasets, inputs, and networks with different random weight initializations.

\subsection{Critical residuals}
\label{ss:transition}
\begin{figure*}[ht]
    \centering
    \includegraphics{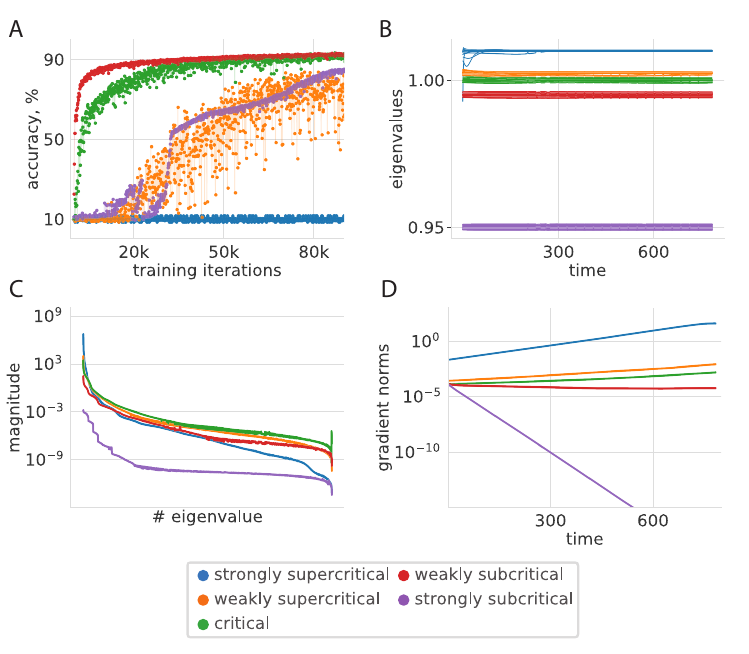}
    \caption{WCRNN performance and dynamics on sMNIST. Colors indicate network type; strongly subcritical ($r=0.95$), weakly subcritical ($r=0.995$), critical ($r=1$), weakly supercritical ($r=1.0025$), strongly supercritical ($r=1.01$).
    \textbf{A}. Test accuracy on sMNIST as a function of training iterations over 200 training epochs. 
    \textbf{B}. Dynamics of eigenvalues of variational term $\mathbf{V}_{x}(f^t)$ before training. Lines show trajectories of 20 randomly chosen eigenvalues over time for a randomly chosen input digit. 
    \textbf{C}. Rank plot of the eigenvalue magnitudes of the Hessian of the loss function $\mathbf{H}_{w}(L)$ before training. Lines show eigenvalues that were computed for a randomly chosen batch of the sMNIST test set.
    \textbf{D}. Evolution of norms of BPTT gradients as a function of time. Lines show gradient norms that were computed over a random input batch before training.}
   \label{fig:grads_all}
\end{figure*}

First, we studied different WCRNNs with network dynamics in proximity of the edge of chaos.
To place networks in this dynamical regime, we introduced linear diagonal residuals of type $\mathbf{R}=r\mathbf{I}$, where the \emph{residual connection strength} $r$ is a scalar hyperparameter, and $\mathbf{I}$ denotes the identity matrix.
According to (\ref{eq:ll_res}), these networks have $N$ Lyapunov exponents with identical values equal to $\log{r}$, where $N$ is the number of units in the network. 
By varying $r$, we can control the Lyapunov exponents, and thereby the distance of the network dynamics to the edge of chaos.
We varied the value of $r$ in the interval $r \in [0.91,1.02]$ to explore the range of subcritical, critical, and supercritical dynamics (see Fig.~\ref{fig:grads_all}). 
To numerically compute Lyapunov exponents from network dynamics, we used a method based on QR-decomposition~\citep{sandri1996numerical} that ensures the numerical convergence of the eigenvalues of the orthonormalized variational term even for unstable dynamics (see \ref{a:algo}).     
Gradient propagation was measured by the $L^{\infty}$ gradient norms $\pdv{L}{f^t}$ in the test dataset~\citep{arjovsky2016unitary}. 
All Hessians were computed for a randomly chosen but fixed set of 1000 samples from the test set and we note that the results were found to be consistent across different choices for this set (data not shown).
The findings were found to be qualitatively consistent in the range $\gamma \in [0.001, 0.1]$ and here we present results for the value $\gamma = 0.01$.
We found that WCRNNs with $\gamma < 0.001$ have difficulties in achieving high task performance as in this case the forcing of the network dynamics by the input becomes too small.
If $\gamma > 0.1$, the forcing of the network dynamics can become too strong and tends to lead to unstable dynamics, which is in agreement with the conditions discussed in Section~\ref{s:resid}.
In particular, the coupling constant $\gamma$ had an effect on learning stability and efficiency, which is similar to the global learning rate, as $\nabla_{w}L= \mathcal{O}(\gamma)$; see Fig.~\ref{fig:a_gamma}.

As expected, we found that the eigenvalues of the variational term had converged to the values defined by the residual according to equation (\ref{eq:ll_res}), which confirms the existence of Lyapunov exponents for the WCRNNs (Fig.~\ref{fig:grads_all}B). 
The learning curves for all networks are shown in Fig.~\ref{fig:a_learn_hess}.
Furthermore, we observed that the magnitudes of the eigenvalues of the Hessians increased with an increase in $r$, see Fig.~\ref{fig:grads_all}C.
This supports the theoretical results on how the proximity of network dynamics of WCRNNs to the edge of chaos affects their effective learning rates, see (\ref{eq:res_contrib}).
In addition, we observed that in subcritical networks the magnitudes of eigenvalues of the Hessian tended to increase over training, while in critical and supercritical networks they tended to decrease; see Fig.~\ref{fig:a_learn_hess}.
The gradient norms $\pdv{L}{f^{t}}$ of the subcritical and supercritical networks were observed to decrease or increase exponentially with a constant rate over time, as predicted by (\ref{eq:res_contrib}), see Fig.~\ref{fig:grads_all}D.
We compare the eigenvalues of the variational term, the eigenvalues of the Hessian, and the gradient propagation before training WCRNNs, because the initial differences between the networks are attributed to the differences in their residuals.
In contrast to the eigenvalues of the Hessian and gradient norms, the eigenvalues of variational term remained in the same range during the training period by design, see Fig.~\ref{fig:a_lls}.

To evaluate the networks' practical expressivity, we measure not only the best accuracy achieved on a test set during the training period (overall performance), but also the learning efficiency.
We assess the learning efficiency by the number of training iterations required for the network to reach a given threshold of minimal performance (MP),   chosen differently for each dataset and thus small values of this measure correspond to better efficiency.
The threshold of minimal performance was set to $50\%$ accuracy for the sMNIST and psMNIST datasets and to a RMS of $0.05$ for the ADD datasets.
These threshold values were chosen to capture a non-negligible deviation from chance-level performance.
For networks that were unable to reach this threshold, the number of iterations to reach minimal performance was set to the maximum number of iterations in the training period.

The practical expressivity of WCRNNs was found to be strongly dependent on the parameter $r$, showing that although residual connections do not limit the theoretical expressivity of the network, they play an important role for network expressivity in practice.
\begin{figure}[ht]
    \centering
    \includegraphics{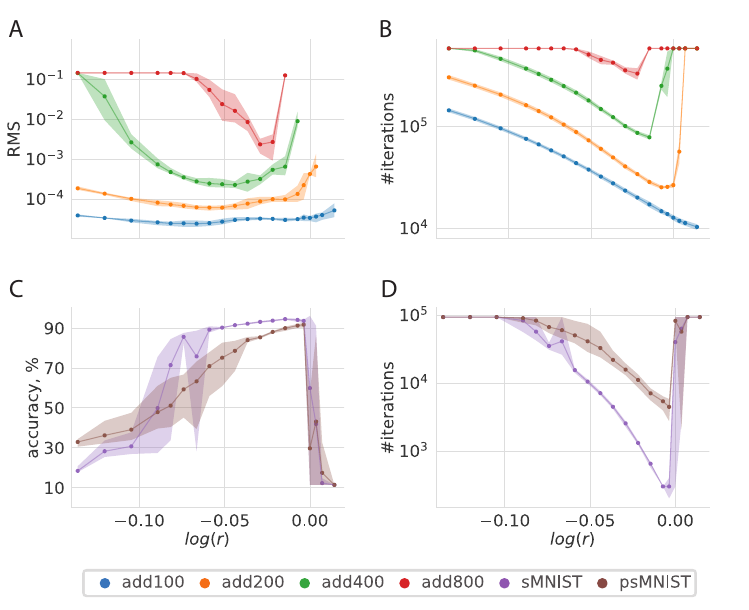} 
    \caption{
    Practical expressivity of WCRNN networks as a function of the value of the residual connection strength $r$ for the ADD and MNIST datasets. 
    Lyapunov exponents of presented WCRNNs are equal to $\log{r}$.
    All networks have a value of $\gamma = 0.01$.
    Lines show mean values over 5 network instances with random weight weight initialization, shaded areas show the range between minimal and maximal values.
    \textbf{A}. Best test accuracy on the ADD task as measured by root mean squared error (RMS) attained over 150 training epochs for the ADD datasets. 
    \textbf{B}. The number of training iterations to reach a defined minimal performance (MP) of 0.05 RMS error (see main text) for ADD datasets. 
    \textbf{C}. Best test accuracy for MNIST dataset over 200 training epochs. 
    \textbf{D}. The number of training iterations to reach a MP of 50\% test accuracy for MNIST datasets.
    }
   \label{fig:perf_speed_all}
\end{figure}

We observed that an increase in distance from the edge of chaos in the subcritical regime resulted in a decrease in learning efficiency for all datasets, see Fig.~\ref{fig:perf_speed_all}.
This is in good agreement with our theoretical predictions presented in Section \ref{s:resid}. 
The observed decay in learning efficiency is caused by vanishing gradients, showing that the EVGP poses an important practical limitation for subcritical networks.
We observed that supercritical networks showed unstable learning trajectories, see Fig.~\ref{fig:perf_speed_all}A, again in line with our theoretical predictions.
The fact that the gradients of supercritical networks were informative but exploded in magnitude was supported by the finding that clipping of the gradients resulted in more stable learning trajectories (data not shown). 
Overall, these results support our hypothesis that proximity to the edge of chaos enables a better learning efficiency at the expense of the stability of learning dynamics.

Interestingly, we observed that supercritical networks showed better performance for the sMNIST and psMNIST datasets, while subcritical networks performed better for the ADD datasets.
This can be explained by the fact that the inputs of ADD datasets are dense in the first dimension (input values in the first dimension are rarely zero), whereas the inputs of the sMNIST datasets are more sparse (many input values are zero or close to it).
These different characteristics of the input favor exploding and vanishing gradients, respectively. 
Importantly, we observed that the best performing networks were closer to the edge of chaos for longer ADD datasets.
This agrees with the general intuition that longer inputs require longer memory timescales and shows that the best inductive biases are dependent on the characteristics of the inputs as defined by the dataset.
We saw that networks with dynamics close to the edge of chaos showed both higher overall performance and better learning efficiency compared to strongly supercritical and strongly subcritical networks. 
Overall, weakly subcritical networks performed best with a dataset-specific optimal distance to the edge of chaos.
We note that our results are in good agreement with previous literature on the role of the interplay between architecture-based inductive biases and characteristic properties of the input~\citep{rajan2010stimulus, mastrogiuseppe2018linking, goyal2022inductive, kerg2022neural, liu2023connectivity}.

In summary, we evaluated the practical expressivity of subcritical, critical, and supercritical WCRNNs by means of their overall performance and learning efficiency on a set of benchmark tasks.
We validated the existence of Lyapunov exponents by numerical calculations and confirmed our theoretical predictions about the trade-off between learning efficiency and the stability of learning dynamics.
Consistent with the previous literature~\citep{schoenholz2016deep}, we found that residual networks with dynamics close to the edge of chaos possess a higher practical expressivity compared to strongly subcritical or supercritical networks.
Importantly, we observed that the weakly subcritical networks showed the best overall performance, and found that the optimal distance to the edge of chaos was indicative of memory timescales beneficial for the task at hand.

\subsection{Rotational residuals}
\label{ss:rot}
The WCRNNs considered so far only had residual matrices with real eigenvalues, so they were limited to scaling linear transformations and reflections. 
To study all geometrical transformations represented by the group of square matrices, we have to introduce rotations, which correspond to matrices with eigenvalues having non-vanishing imaginary part. 
For this purpose, we consider residual matrices $\mathbf{R}$ taking the form of orthonormal diagonal block matrices
\begin{equation}
\mathbf{R}
 =
  \begin{bmatrix}
   \mathbf{T} _{1} &
  \mathbf{0} &
  \cdots &
   \mathbf{0}  \\
   \mathbf{0}  &
  \mathbf{T}_{2} &
  \cdots &
   \mathbf{0}   \\
   \vdots &
   \vdots &
   \ddots &
   \vdots  \\
   \mathbf{0}  &
   \mathbf{0}  &
    \cdots &
  \mathbf{T}_{N/2}  \\
   \end{bmatrix},
   \quad\text{with}\quad 
   \mathbf{T}_{i}
 =
  \begin{bmatrix}
   \cos{\phi_{i}} &
  -\sin{\phi_{i}}\\
   \sin{\phi_{i}}  &
  \cos{\phi_{i}}
   \end{bmatrix},
   \label{eq:rot}
\end{equation}
where $N$ is the number of units in the network
and $\phi_{i} \in [0,2\pi]$ denotes an angular frequency of rotation.
This type of residual represents rotation matrices with arbitrary combinations of angular frequencies $\phi_{i}$. 

\begin{figure}
    \centering
    \includegraphics{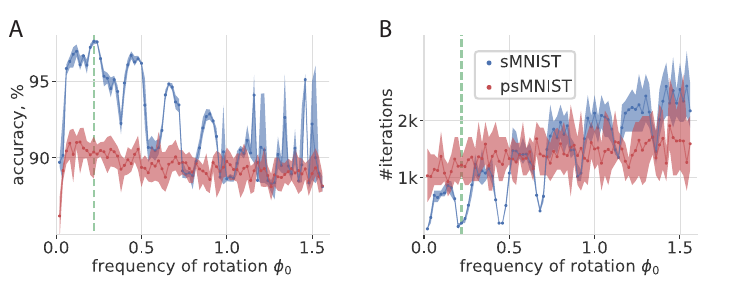}
    \caption{
    Performance of WCRNNs with homogeneous rotational residuals on sMNIST and psMNIST. Lines show mean values over 5 network instances with random weight initialization, shaded areas show the range between maximal and minimal values. Colors indicate the training dataset. The green dashed line indicates the characteristic frequency of sMNIST $\phi_c = 2\pi/28 \approx 0.22$. 
    \textbf{A}. The best test accuracy attained as a function of angular frequency of rotation $\phi_0$ of the homogeneous residual matrix. 
    \textbf{B}. The number of training iterations to reach a defined minimal performance (MP) of 50\% test accuracy (see main text), as a function of angular frequency of rotation $\phi_0$.
    }
   \label{fig:rot}
\end{figure}
First, we studied WCRNNs with rotational residuals of the form (\ref{eq:rot}) for which all $\phi_{i}$ have a constant value $\phi_0$, referred to as \emph{homogeneous rotational residuals} from now on.
We trained WCRNNs with homogeneous rotational residuals for different values of $\phi_0$ on sMNIST and psMNIST for 50 epochs, see Fig.~\ref{fig:rot}.
Here, we chose a shorter training period to better demonstrate the differences in performance between such networks that arise due to differences in their learning efficiency, but we also trained these networks for 200 epochs, see supplementary Fig.~\ref{fig:a_rot} where the results were similar.
We found that these networks converged fastest and expressed highest performance on sMNIST when their angular frequency $\phi_0$ was close to $2\pi/28 \approx 0.22$.
Interestingly, this frequency coincides with the maximal peak of the average power spectra of the samples in sMNIST, thus we call it the \emph{characteristic frequency} $\phi_c$ of the dataset~\cite{effenberger2022biology}. 
We found that homogeneous rotational residuals with angular frequencies that were close to integer multiplies of $\phi_c$ (i.e., $\phi_c$ and its harmonics) result in networks that show the highest performance and the best learning efficiency, see Fig.~\ref{fig:rot}.
As follows from Section \ref{ss:learning}, we attribute this enhanced performance to the fact that the instantaneous Jacobians of the networks with rotational residuals modulate the effective learning rates in an oscillatory manner.
Such instantaneous Jacobians allow the networks to accumulate derivatives that are in-phase and cancel out derivatives that are out of phase, see Fig.~\ref{fig:a_rot}B. 
Taken together, this shows that oscillatory learning rates can be beneficial for a network's practical expressivity when they align with characteristic spectral properties of the dataset. 

In contrast, we did not find a strong effect on practical expressivity when varying the angular frequencies of the rotational residuals for the psMNIST, see Fig.~\ref{fig:rot}.
This can be explained by the fact that, in contrast to the sMNIST dataset, the samples of the psMNIST dataset do not possess a prominent characteristic frequency that can be exploited for classification.

Furthermore, we also studied the case where residuals are given by random orthonormal matrices with a uniform distribution of angular frequencies of rotations (constructed with the function \texttt{scipy.stats.ortho\_group.rvs} from the SciPy package~\citep{2020SciPy-NMeth}).
The latter were found to result in networks with a performance similar to that of those with residuals with heterogeneous angular frequencies discussed below, see Fig.~\ref{fig:a_rot}D.

Overall, we found that choosing residuals informed by characteristic spectral properties of the samples in the dataset can result in WCRNNs with higher practical expressivity.
In particular, we show for sMNIST that rotational residuals can significantly improve the efficiency of learning. 
This agrees with the previous findings of increased performance of networks composed of oscillatory units~\citep{norcliffe2020second, rusch2020coupled,effenberger2022biology}.

\subsection{Heterogeneous residuals}
\label{ss:hete}
\begin{figure}[ht]
    \centering
    \includegraphics{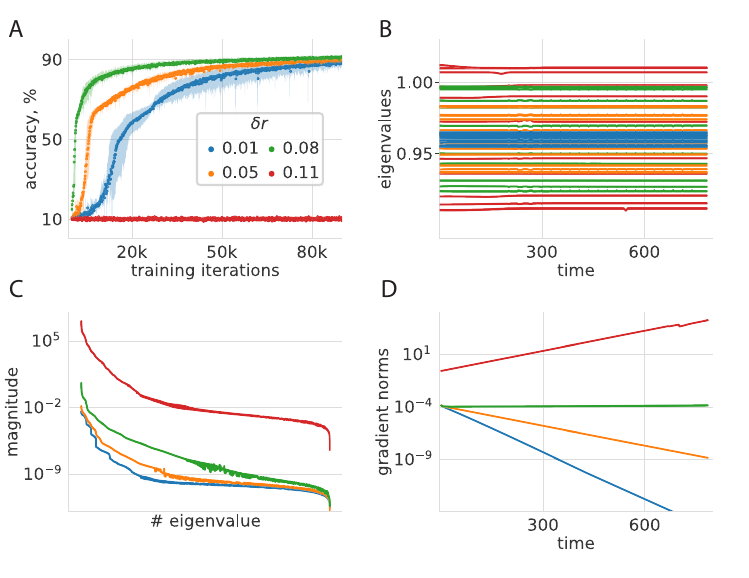} 
    \caption{
    Performance of heterogeneous subcritical WCRNNs on sMNIST with $r_0=0.96$ and different levels of heterogeneity $\delta r$ (color-coded). 
    \textbf{A}. Test accuracy as a function of training iterations over 200 training epochs. The lines show mean test accuracy over 5 network instances with random weight initialization, shaded areas show the range between minimal and maximal values of test accuracy. 
    \textbf{B}. Dynamics of eigenvalues of variational term $\mathbf{V}_{x}(f^t)$ for WCRNNs with different levels of heterogeneity. All eigenvalues computed using the same randomly chosen input. Lines show trajectories of 15 randomly chosen eigenvalues over time. 
    \textbf{C}. Rank plot of the eigenvalue magnitudes of the Hessian of the loss function $\mathbf{H}_{w}(L)$ before training. Lines show eigenvalues that were computed for a randomly chosen batch on sMNIST test dataset. 
    \textbf{D}. Evolution of norms of BPTT gradients as a function of time. Lines show gradient norms that were computed over a random input batch before training.}
    \label{fig:hete_crit}
\end{figure}

In this section, we study how introducing heterogeneous residuals influences practical expressivity in WCRNNs. 
We first considered networks for which the residual matrix takes the form 
\begin{equation}
\mathbf{R}=\operatorname{diag}(\mathbf{r}),
\label{eq:het}
\end{equation}
with each coordinate $r_{i}$ of $\mathbf{r}$ sampled from a uniform distribution $\operatorname{U}(r_{0}-\delta r/2,r_{0}+\delta r/2)$. 
For our experiments, we fix $r_{0}$ which defines a baseline distance to edge of chaos and gradually increase the level of heterogeneity controlled by the term $\delta r$.
We found that subcritical networks $r_{0}<1$ with moderate heterogeneity ($\delta r/2$ is less than the distance to the edge of chaos) showed better performance and efficiency of learning, and only strong heterogeneity ($\delta r/2$ is greater than the distance to the edge of chaos) yield WCRNNs with unstable learning dynamics, see Fig.~\ref{fig:hete_crit}A.
This can be explained by the fact that the heterogeneity of the residual matrix increased the diversity of memory timescales in the Lyapunov spectrum and brought the network dynamics closer to the edge of chaos, resulting in longer memory timescales, see Fig.~\ref{fig:hete_crit}B.
Moreover, the heterogeneity in subcritical networks improved gradient propagation and increased the number of non-vanishing eigenvalues of the Hessian, making learning dynamics richer, see Fig.~\ref{fig:hete_crit}C and D.
In contrast, for networks with dynamics close to the edge of chaos, the benefit of having heterogeneous residuals reduced, because heterogeneity increased the risk of exploding gradients (data not shown).

\begin{table}
  \caption{Practical expressivity of informed heterogeneous WCRNNs compared to homogeneous WCRNNs, assessed by best test accuracy and the number of training iterations to reach a minimal performance (MP).}
  \label{t:het}
  \centering
  \begin{tabular}{ p{2cm} p{3cm} p{3cm} p{3cm} p{3cm}  }
    \toprule
     & Homogeneous
    & Heterogeneous
    & Homogeneous
    & Heterogeneous \\ [0.5ex]
    \midrule
    &accuracy, \% 
    &accuracy, \%
    &iter. to MP
    &iter. to MP \\ [0.5ex]
    sMNIST & 96.05   & 98.22 & 200 & 200  \\
    spMNIST  & 92.66  & 95.37 & 2076 & 1038  \\
    \midrule
    & RMS 
    & RMS
    &iter. to MP
    &iter. to MP \\ [0.5ex]
    ADD100 & $2.0 \cdot 10 ^ {-5}$   & $4.6 \cdot 10^{-5}$ & 9298 & 8698    \\
    ADD200  & $5.1 \cdot 10^{-5}$   & $3.3 \cdot 10^{-5}$ & $2.4 \cdot 10^{4}$ & $3.1 \cdot 10^{4}$     \\
    ADD400  & $1.5 \cdot 10^{-4}$   & $3.1 \cdot 10^{-4}$ & $7.4 \cdot 10^{4}$ & $9.5 \cdot 10^{4}$   \\
    ADD800  & $9.1 \cdot 10^{-4}$   & $19.1 \cdot 10^{-4}$ & $2.9 \cdot 10^{5}$ & $4.1 \cdot 10^{5}$   \\
    \bottomrule
  \end{tabular}
\end{table}
Next, we considered WCRNNs with residual matrices that are a product of an orthonormal matrix of the form (\ref{eq:rot}) and a diagonal matrix of the form (\ref{eq:het}).
Furthermore, we allowed unit-specific heterogeneity of the weak coupling parameter by substituting the scalar coupling constant from (\ref{eq:general_residual}) by a vector $\boldsymbol{\gamma}$.
Informed by previous experiments, we sampled $\mathbf{r}_{i}$ from $U([0.99,1])$, $\phi_{i}$ from $U([0,\pi/4])$, and $\boldsymbol{\gamma}_i$ from $U([0.005, 0.05])$ independently for each network unit.
We found that WCRNNs with such informed heterogeneity performed on par with the best homogeneous configuration of WCRNNs on all datasets considered, see Table \ref{t:het}.
Our results show that the variety of memory timescales present in WCRNNs with heterogeneous residuals allows them to generalize well over different datasets and therefore obtain increased practical expressivity compared to networks with homogeneous residuals.
In particular, it follows that informed heterogeneity can be used to avoid a computationally expensive search for the best performing residual configuration.

\subsection{Non-linear residuals}
\label{ss:init}
So far, we have only studied WCRNNs for which the residual was given by a linear map.
In this section, we study two non-linear variants of WCRNNs of the form
\begin{equation}
\mathbf{x}_{t+1} =  \sigma(\mathbf{R}\mathbf{x}_t)  + \gamma \sigma(\mathbf{W}\mathbf{x}_t + \mathbf{S}_t).
\label{eq:non_res}
\end{equation}
and
\begin{equation}
\mathbf{x}_{t+1} =  \sigma(\mathbf{R}\mathbf{x}_t  + \gamma (\mathbf{W}\mathbf{x}_t + \mathbf{S}_t)).
\label{eq:weak_init}
\end{equation}
For ease of notation, we will refer to networks defined by (\ref{eq:non_res}) as type A and to networks defined by (\ref{eq:weak_init}) as type B.
Note that the observations from Section \ref{s:resid} also hold for the non-linear case presented here, meaning that the weak coupling ensures that the memory properties of the networks (\ref{eq:non_res}) and (\ref{eq:weak_init}) are still mostly determined by the residual connections.
Moreover, when the non-linearity is not in a strongly saturating regime, the eigenvalues of the residuals matrices of both network types A, B still influence the network dynamics to a large extent. 
This is why we continue to distinguish subcritical, critical, and supercritical nonlinear WCRNNs,as before.
We note that due to the presence of the non-linearity $\sigma$, this classification is less strict than for WCRNNs with linear residuals. 
To investigate these networks, we performed the same set of experiments on the MNIST datasets as in Sections~\ref{ss:transition}, \ref{ss:rot}, and \ref{ss:hete}. 

\begin{figure}[ht]
    \centering
    \includegraphics{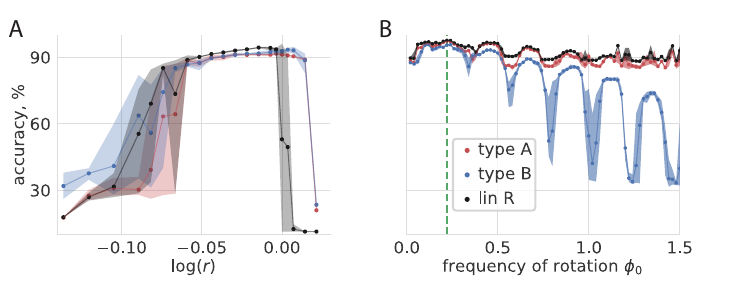}
    \caption{Performance of WCRNNs with non-linear residuals of type A and type B on sMNIST in comparison with their linear analogs from previous experiments. The lines show the mean test accuracy over 5 network instances with random weight initialization, shaded areas show the range between maximal and minimal values of the test accuracy. 
    \textbf{A}. The best test accuracy as a function of the maximal eigenvalue of the residual matrix (as controlled by the scalar parameter $r$). 
    \textbf{B}. The best test accuracy as a function of angular frequency of rotation $\phi_0$ of the homogeneous residual matrix. The green dashed line indicates the characteristic frequency of sMNIST $\phi_c = 2\pi/28 \approx 0.22$}
   \label{fig:weak_init}
\end{figure} 
When training networks of both types, we found the same trade-off between efficiency and stability of learning dynamics as observed in networks with linear residuals.
Similarly to linear networks, the non-linear WCRNNs with the highest performance and best learning efficiency were the networks that have eigenvalues of residual matrices close to critical value $1$, see Fig.~\ref{fig:weak_init} A.
Subcritical networks of type A showed the best performance and the fastest convergence, in contrast to type B networks, for which the best practical expressivity was achieved by weakly supercritical networks.
For both studied non-linear variants of WCRNNs, we also found peaks in the learning efficiency and performance when trained with rotational residuals on sMNIST, see in Fig.~\ref{fig:weak_init} B.
Interestingly, we found that the performance of networks of type B showed stronger sensitivity to angular frequencies of rotational residuals compared to networks of type A.
These differences can be explained by the fact that one common non-linearity better prevents chaotic dynamics, but makes biases from initialization have a stronger effect on performance.
We also observed that both network types were able to achieve similar performance levels as their highest performing homogeneous variants when equipped with informed heterogeneity from Section \ref{ss:hete} (data not shown).

Notably, the network of type B can be considered as a weakly coupled residual initialization of an Elman RNN, because the residual matrix and other weight matrices are subject to the same non-linearity.
In agreement with previous studies~\citep{arjovsky2016unitary}, learning long tasks was found to be challenging for classic Elman networks even after applying standard techniques such as identity residuals and gradient clipping (Table \ref{t:compare}, $\text{Elman}_{\text{res+clip}}$ and $\text{Elman}_{\text{wc}}$), emphasizing the advantage of the weakly coupled residual initialization scheme presented in (\ref{eq:weak_init}).

We furthermore compared the presented variants of WCRNNs with other architectures that are known to be well-suited for working with long sequences such as LSTM~\citep{gu2020improving} and S4~\citep{gu2021efficiently}, see Table~\ref{t:compare}.
This comparison also includes the more challenging grayscale sequential CIFAR dataset (sCIFAR10) which is part of Long Range Arena benchmark designed to evaluate model performance on long sequences~\citep{tay2020long}. 
We also compared these networks with a variant of WCRNN that lacks the non-linearity ($\text{WCRNN}_{\text{no tanh}}$), showing the importance of the non-linear activation function. 
We found that heterogeneous WCRNNs with informed memory properties are able to achieve performance levels that are competitive with the much bigger and complex SOTA models for some datasets. 

\begin{table}
\caption{Best performance of various RNN architectures on a number of benchmark tasks. $\text{Elman}_{\text{res+clip}}$ denotes the Elman network with an identity residual and gradient clipping. $\text{Elman}_{\text{wc}}$ denotes the Elman network with weakly coupled residual initialization (type B network). $\text{WCRNN}_{\text{best}}$ denotes the informed heterogeneous WCRNN. $\text{WCRNN}_{\text{no tanh}}$ denotes a WCRNN with the same residual and $\gamma$ as $\text{WCRNN}_{\text{best}}$, but without the tanh non-linearity. LSTM denotes the default PyTorch implementation of an LSTM network. 
Elman and WCRNN networks have 100 units, LSTM networks 47 units (resulting in each model having around 10k trainable parameters). State-of-the-art results are given in the form of UR-LSTM (around 250k parameters for ADD, around 4M parameters for sMNIST/psMNIST, around 32M parameters for sCIFAR10) and S4 networks (around 300k parameters).
Note the significantly increased model sizes of UR-LSTM and S4 in with respect to the other models.
A dash (-) indicates that no data is available.
  }
  \label{t:compare}
  \centering
  \begin{tabular}{ p{5.0cm} p{1.6cm} p{1.6cm} p{4.5cm} p{1.6cm} }
    \toprule
    & sMNIST
     & psMNIST
     & ADD100, 200, 400, 800
    & sCIFAR10 \\ [0.1ex]
    &acc. \% 
    &acc. \% 
    &RMS
    &acc. \% \\ [0.5ex]
    \midrule
    $\text{Elman}_{\text{res+clip}}$ & $36.53$ & $78.59$ & $1.5 \cdot 10^{-5}$, $3.8 \cdot 10^{-5}$, fail, fail & $18.7$\\ [0.1ex]

    $\text{Elman}_{\text{wc}}$  & $97.33$ & $92.98$  & $1.1 \cdot 10^{-3}$, $2.9 \cdot 10^{-3}$, $8.0 \cdot 10^{-2}$, fail & $38.49$ \\ [0.1ex]

    $\text{WCRNN}_{\text{best}}$ & $98.22$ & $95.37$   & $4.6 \cdot 10^{-5}$, $3.3 \cdot 10^{-5},$ $3.1 \cdot 10^{-4}$, $1.9 \cdot 10^{-3}$  & $47.21$  \\ [0.1ex]

    $\text{WCRNN}_{\text{no tanh}}$ & $91.15$ & $90.11$  & fail, fail, fail, fail & $29.79$ \\ [0.1ex]

    LSTM  & $93.3$ & $90.29$ & $1.0 \cdot 10^{-6}$, $1.0 \cdot 10^{-6}$, $2.0 \cdot 10^{-6}$, $6.0 \cdot 10^{-6}$ & $58.13$ \\ [0.1ex]

    UR-LSTM~\citep{gu2020improving}   & $99.28$ & $96.9$6 & $1 \cdot 10^{-10}$ (ADD2000) & $71.0$ \\ [0.1ex]

    S4~\citep{gu2021efficiently}    & $99.63$ & $98.7$ & -  & $91.8$ \\ [0.1ex]
    \bottomrule
  \end{tabular}
\end{table}

\section{Discussion}
\noindent

In this work, we introduced weakly coupled residual recurrent networks (WCRNNs) and studied how their recurrent residual connections influence network dynamics, properties of fading memory, and practical expressivity.
A dynamical systems analysis of WCRNNs allowed us to uncover a connection between network dynamics and the weight dynamics resulting from BPTT training.
Based on this analysis, we predicted a trade-off between the stability of the learning dynamics and the learning efficiency of the backpropagation algorithm.
In line with this prediction, simulation results showed that WCRNNs with dynamics close to the edge of chaos achieve greater practical expressivity than more subcritical or supercritical networks.
Moreover, we found that several classes of informed residual connections could yield effective inductive biases for WCRNNs.
In particular, we found that (i) rotational residuals are beneficial when they match the characteristic spectral properties of the data, (ii) residuals resulting in subcritical fading memory are favorable when temporally distant dependencies are present in the data, and (iii) heterogeneous residuals can increase the networks' practical expressivity by providing an informed range of memory timescales.
In addition, heterogeneity can help avoid the computationally expensive search required to find the best-performing configuration for a homogeneous network.

Over the years, many approaches have been proposed to overcome the EVGP encountered when training RNNs with BPTT.
Among those are gated architectures, such as long-short-term memory networks (LSTM)~\citep{hochreiter1997long} and gated recurrent units (GRU)~\citep{cho2014learning}, which in many cases still require gradient clipping to achieve high practical expressivity~\citep{pascanu2013difficulty}. 
Furthermore, models that place constraints on weight matrices such as orthogonal~\citep{helfrich2018orthogonal}, unitary~\citep{arjovsky2016unitary} or antisymmetric weight matrices~\citep{chang2019antisymmetricRNN} were proposed to mitigate the EVGP.
In contrast to these models, the WCRNNs proposed here do not impose strict conditions on weight matrices, as this can limit the practical expressivity of networks~\citep{vorontsov2017orthogonality}. 
In that sense, WCRNNs are similar to RNNs with a specific initialization of their recurrent weights (for example, using the identity or orthogonal matrices~\citep{le2015simple,mishkin2015all}), but differ in that the weak coupling in WCRNNs ensures the stability of the gradient properties throughout the learning process.

We also note that in our experiments we only used normal residual matrices and that it is known that non-normal initialization can lead to higher levels of practical expressivity~\cite{kerg2019non}. 
Thus, the thorough study of how normality, diagonalizability, and other properties of residual matrices affect the performance of WCRNNs could be a potential direction for future research. 

Although residual networks were first introduced without dynamical systems theory
in mind, it was later shown that residual networks possess efficient gradient propagation properties in the infinite-width approximation when their dynamics are close to the edge of chaos ~\citep{yang2017mean}. 
Moreover, this approximation has been used to show the role of initialization schemes in shaping ``lazy'' or ``rich'' regimes of learning dynamics ~\cite{chizat2019lazy,geiger2020disentangling,flesch2021rich,liu2023connectivity}.
Interestingly, the presence of weak coupling in WCRNNs resembles the infinite-width approximation as the influence of recurrent weights is scaled down by the weak coupling factor $\gamma$.
We found that over training, the weights of WCRNNs often changed by an order of magnitude with respect to their initialization values, indicating a ``rich'' learning regime.
However, the unique property of WCRNN is that the weak coupling allows for ``rich'' learning while maintaining stable memory properties.

Residual networks also sparked new interest in a dynamical systems approach due to the fact that in the limit of an infinite-depth approximation they can be understood as a system of differential equations, the NeuralODE approach~\citep{chen2018neural}.
However, in contrast to WCRNNs, architectures based on the NeuralODE approach are limited to describe continuous-like dynamics.
Notably, recent research on various weight initialization methods in NeuralODE approach has revealed the crucial role of the properties of network dynamics for the training of these networks \cite{jarne2023different,christodoulou2022regimes,jarne2023exploring}, in line with the findings of the present study.

Furthermore, recently introduced methods grounded in geometric principles and similarity metrics~\cite{ostrow2023beyond, schuessler2023aligned} allow for the detection of dynamical structures in recurrent neural networks.
The application of these methods to WCRNNs is left for a future study.

Our results are consistent with previous studies on the benefits of residual networks with dynamics at the edge of chaos~\citep{schoenholz2016deep}, and also provide an additional perspective on the previously shown computational advantage of recurrent networks consisting of oscillatory units over non-oscillating architectures~\citep{norcliffe2020second, rusch2020coupled, effenberger2022biology}. 
Our findings also agree with studies in the field of neuroscience, indicating that the brain seems to operate close to criticality but in a slightly subcritical regime, as this has computational advantages~\citep{wilting2018operating, wilting201925}.
In addition, our results also agree with recent studies that suggest the functional role of neural heterogeneity~\citep{perez2021neural, effenberger2022biology,sanchez2023heterogeneity}.

We also anticipate that our results will be relevant in the context of continuous learning, where effective memory is known to be essential to avoid catastrophic forgetting~\citep{hadsell2020embracing}.
Furthermore, we hypothesize that our approach to incorporate informed biases into residuals could find its application in feedforward architectures, because residual connections are widely used in a range of modern architectures~\citep{ronneberger2015u,he2016deep,vaswani2017attention}.

\section{Conclusion}

\noindent

In a broader context, we believe that the findings presented here are in line with a recent surge in interest in RNNs, seeking to overcome the EVGP and training inefficiencies~\citep{orvieto2023resurrecting,zucchet2023online}.
These developments show that the careful design of RNNs can result in state-of-art performance on long range memory tasks.
In some cases, RNNs were shown to surpass the performance of feed-forward Transformer-based architectures~\citep{tay2022efficient}, while overcoming the drawbacks of their dot-product attention, where memory and computational complexity exhibit quadratic scaling with sequence length~\cite{peng2023rwkv}.
In this work, we have defined weakly coupled recurrent networks (WCRNNs) that possess well-defined Lyapunov exponents and have shown how the practical expressivity and training stability of WCRNNs are influenced by their dynamics, and how heterogeneous and informed residuals can increase further practical expressivity without increasing system size.
We believe that this is an important step forward in the understanding of residual RNNs on both a theoretical and practical level. 

In summary, our results show how the properties of fading memory resulting from RNN dynamics play a crucial role in shaping the backpropagation-induced learning dynamics, and how the implementation of informed memory properties by means of residual connections can improve the practical expressivity of RNNs.

\bibliographystyle{elsarticle-harv} 
\bibliography{references}

\appendix
\clearpage

\section{Algorithm for computation of Lyapunov exponents}
\label{a:algo}

Here, we explain how we compute the Lyapunov exponents for WCRNNs defined by equation (\ref{eq:disc_map}). In theory, Laypunov exponents can be computed directly from the Oseledts equation (\ref{eq:ose}), but in practice the $P$-dimensional volume of the tangent space tends to align with the eigenvector associated with the largest eigenvalue, resulting in degenerate matrices and therefore numerical problems. To overcome such problems, an orthonormalization procedure was introduced; see, for example,~\citep{sandri1996numerical}. This procedure orthonormalizes the tangent space along a system trajectory that gives more stable estimates of the scaling of the $P$ dimensional volume. For the case of a discrete system (\ref{eq:disc_map}), we iterate it alongside with its variational equation
\begin{equation}
\delta \mathbf{P}(t+1) = \mathbf{J}_{\mathbf{x}}(f^t(\mathbf{x}_1)) \delta \mathbf{P}(t).
\label{eq:a_var_general}
\end{equation}
If we apply the chain rule for $t$ times, we obtain Equation (\ref{eq:var_general_vol}) from the main text

\begin{equation}
\delta \mathbf{P}(t+1) = \mathbf{V}_{\mathbf{x}}(f^t(\mathbf{x}_1)) \delta \mathbf{P}(1), 
\label{eq:a_var_main_text}
\end{equation}

where $\mathbf{V}_{\mathbf{x}}(f^t(x_1)) = \mathbf{J}_{\mathbf{x}}(f^{t}(x_1))\mathbf{J}_{\mathbf{x}}(f^{t-1}(\mathbf{x}_1))...\mathbf{J}_{\mathbf{x}}(f(\mathbf{x}_1))$. To compute the Lyapunov exponents, we need to estimate the eigenvalues of the variational term $\mathbf{V}_{\mathbf{x}}(f^t(x_1))$ with the initial condition of $\delta \mathbf{P}(1)=\mathbf{I}$.  

The orthonormalization procedure is performed as follows. After every iteration of the system (\ref{eq:disc_map}), we compute its Jacobian and evolve the variational equation (\ref{eq:a_var_general}), where the $P$ dimensional volume of the tangent space is initialized with the identity matrix. Next, we perform a QR decomposition on the resulting volume of the tangent space $\mathbf{Q}_1$. The QR decomposition of $\mathbf{Q}_1$ produces two matrices $\mathbf{Q}_2$ and $\mathbf{R}_2$, where the orthonormal matrix $\mathbf{Q}_2$ defines the rotational component, and the diagonal elements of $\mathbf{R}_2$ define the volume scaling. Finally, we collect the eigenvalues of $\mathbf{R}$, which describe the instantaneous transformation of the volume, and perform the next step with a new volume from the tangent space $\mathbf{Q}_2$. Furthermore, if we repeat this procedure for $t$ times, the final variational term is 

\begin{equation}
\mathbf{V}_{\mathbf{x}}(f^t(\mathbf{x}_1)) =  \mathbf{Q}_t\mathbf{R}_t \dots \mathbf{R}_1 .
\label{eq:a_normalization}
\end{equation}

From (\ref{eq:a_normalization}) it follows that if the logarithms of the eigenvalues $\mathbf{R}_t$ converge, they define unique Lyapunov exponents according to the Oseledets equation (\ref{eq:ose}). Therefore, we show the convergence of eigenvalues of the variational term in all of our figures. We also want to note that this method is usually applied to autonomous systems, but the design of the residuals of the system (\ref{eq:general_residual}) allows us to study non-autonomous dynamics. 

The pseudocode for the orthonormalization algorithm is as follows:

\begin{algorithm}
\caption{Collect eigenvalues of the variational term}
\begin{algorithmic} 
\STATE \textbf{Initialize:} $\mathbf{x}, \mathbf{Q}, \mathbf{I}$
\WHILE{$t \leq L$}
\STATE $\mathbf{x} \leftarrow \mathbf{f}(\mathbf{x},\mathbf{I})$
\STATE $\mathbf{J} \leftarrow \dv{\mathbf{f}}{\mathbf{x}}$
\STATE $\mathbf{Q} \leftarrow \mathbf{J}\mathbf{Q}$
\STATE $\mathbf{Q},\mathbf{R} \leftarrow qr(\mathbf{Q})$
\STATE $\lambda \gets \text{append}(\operatorname{diag}(R))$
\ENDWHILE
\end{algorithmic}
\end{algorithm}

\clearpage
\section{Supplementary figures}
\label{a:supp_figs}
\setcounter{figure}{0}

\begin{figure}[ht]
    \centering
    \includegraphics{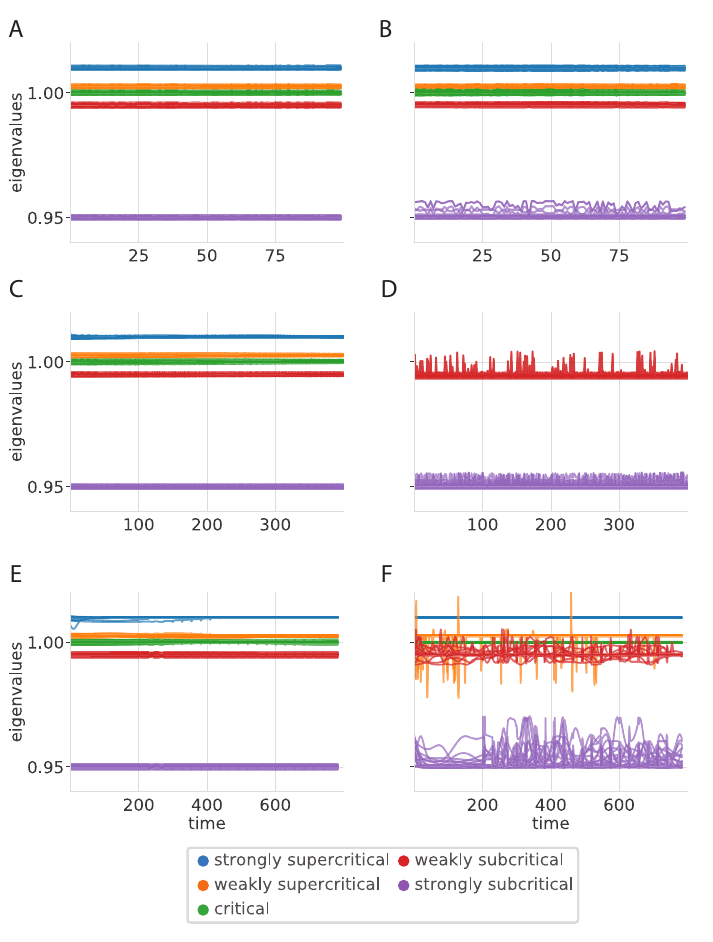}
    \caption{
    Dynamics of eigenvalues of variational term $\mathbf{V}_{x}(f^t)$ before training and after 200 training epochs. Lines show trajectories of 20 randomly chosen eigenvalues over time for a randomly chosen input digit. Colors indicate network type, strongly supercritical have $r=1.01$, weakly supercritical have $r=1.0025$, critical have $r=1$, weakly subcritical have $r=0.995$, strongly subcritical have $r=0.95$.
    A. ADD100 before training; 
    B. ADD100 after training; 
    C. ADD400 before training; 
    D. ADD400 after training;
    E. psMNIST before training; 
    F. psMNIST after training;}
   \label{fig:a_lls}
\end{figure}
\begin{figure}[ht]
    \centering
    \includegraphics{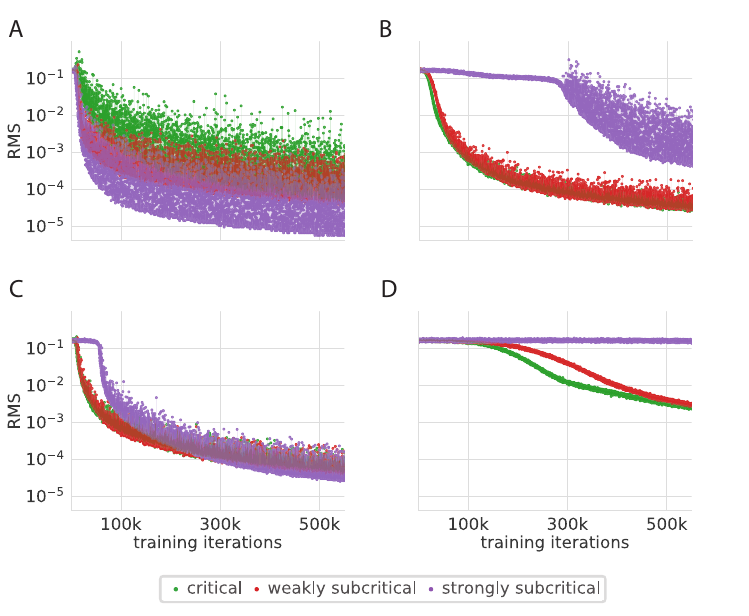}
    \caption{
    Learning trajectories for WCRNNs subject to different learning rates and coupling constants $\gamma$, trained on the ADD100 dataset. Lines show test accuracy measured in RMS as a function of training iterations over 150 training epochs. 
    A. Learning rate: $\eta=0.1$, coupling constant: $\gamma=0.01$. Note that the learning dynamics are unstable.
    B. Learning rate: $\eta=0.1$, coupling constant: $\gamma=0.001$. Note that the decrease in $\gamma$ results in more stable but slower learning dynamics compared to A.
    C. Learning rate: $\eta=0.01$, coupling constant: $\gamma=0.01$. Note that the decrease in learning rate results in more stable but also slower learning dynamics.
    D. Learning rate: $\eta=0.01$, coupling constant: $\gamma=0.001$. Note the very stable and also very slow learning dynamics.}
   \label{fig:a_gamma}
\end{figure}

\begin{figure}[ht]
    \centering
    \includegraphics{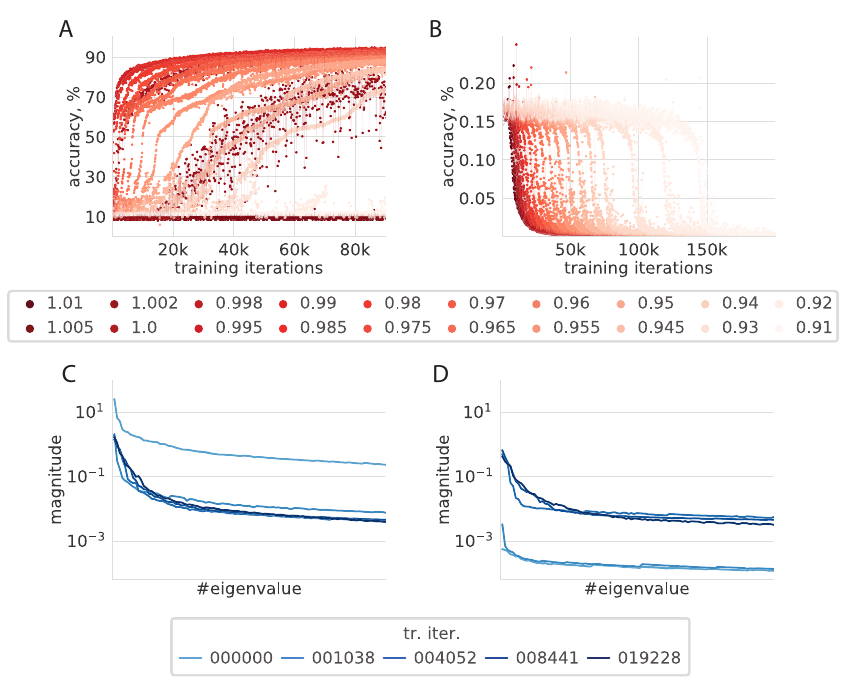}
    \caption{
    Learning dynamics for WCRNNs. Plot A and B show learning trajectories for the studied range of the residual connection strength $r \in [0.91,1.02]$. Rank plots C and D show the eigenvalue magnitudes of the Hessian of the loss function $\mathbf{H}_{w}(L)$ during training. The eigenvalues were computed for a randomly chosen batch from the sMNIST test set.
    A. Lines show test accuracy as a function of training iterations for sMNIST dataset. 
    B. Lines show test error measured in RMS as a function of training iterations for the ADD100 dataset.
    C. Lines show eigenvalues for a critical WCRNN. Note the decrease in magnitudes over learning.
    D. Lines show the eigenvalues for a strongly subcritical WCRNN. Note the increase in magnitudes over learning.
    }
   \label{fig:a_learn_hess}
\end{figure}

\begin{figure}[ht]
    \centering
    \includegraphics{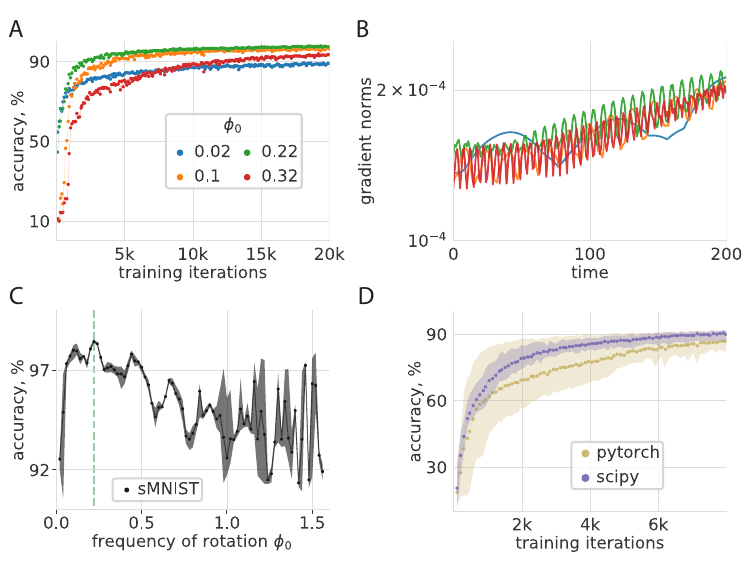}
    \caption{
    A. Learning trajectories for WCRNNs with different angular frequencies of the rotational residuals $\phi_0$, trained on the sMNIST dataset. Lines show the test accuracy as a function of training iterations.
    B. Evolution of the norms of the gradients $\pdv{L}{f^t}$ as a function of inference time for homogeneous WCRNNs with different angular frequencies of the rotational residuals $\phi_0$. Lines show gradient norms computed on a random batch of the sMNIST test set before training.
    C. Best test accuracy of rotational WCRNNs trained on sMNIST over 200 epochs as a function of the angular frequency of the rotational residual $\phi_0$.
    D. Comparison between different implementations for the construction of heterogeneous orthonormal residual matrices, comparing \texttt{scipy.stats.ortho\_group.rvs} implementation (red) and our \texttt{pytorch.rand} implementation (blue). Test accuracy for WCRNNs with different residual matrices as a function of training iterations for first 7000 training iterations. Lines show average accuracy over 5 networks with random weight initialization, shaded area indicates the range between minimal and maximal values.
    }
   \label{fig:a_rot}
\end{figure}

\end{document}